\documentclass[10pt,twocolumn,letterpaper]{article}

\usepackage{cvpr}
\usepackage{times}
\usepackage{epsfig}
\usepackage{graphicx}
\usepackage{amsmath}
\usepackage{amssymb}
\usepackage{subfigure}
\usepackage{tabularx}
\usepackage{arydshln}
\usepackage{multirow}
\usepackage{array}
\usepackage{bm}
\usepackage{soul}
\usepackage[export]{adjustbox}

\newcolumntype{C}[1]{>{\centering\let\newline\\\arraybackslash\hspace{0pt}}p{#1}}

\usepackage[pagebackref=true,breaklinks=true,letterpaper=true,colorlinks,bookmarks=false]{hyperref}

\cvprfinalcopy 

\def\cvprPaperID{1855} 

\ifcvprfinal\fi
\begin{document}

\title{Image Question Answering using Convolutional Neural Network \\ with Dynamic Parameter Prediction}

\author{Hyeonwoo Noh\hspace{1.7cm}Paul Hongsuck Seo\hspace{1.7cm}Bohyung Han\\
Department of Computer Science and Engineering, POSTECH, Korea\\
{\tt\small \{hyeonwoonoh\_, hsseo, bhhan\}@postech.ac.kr}}

\maketitle

\begin{abstract}
We tackle image question answering (ImageQA) problem by learning a convolutional neural network (CNN) with a dynamic parameter layer whose weights are determined adaptively based on questions.
For the adaptive parameter prediction, we employ a separate parameter prediction network, which consists of gated recurrent unit (GRU) taking a question as its input and a fully-connected layer generating a set of candidate weights as its output.
However, it is challenging to construct a parameter prediction network for a large number of parameters in the fully-connected dynamic parameter layer of the CNN.
We reduce the complexity of this problem by incorporating a hashing technique, where the candidate weights given by the parameter prediction network are selected using a predefined hash function to determine individual weights in the dynamic parameter layer.
The proposed network---joint network with the CNN for ImageQA and the parameter prediction network---is trained end-to-end through back-propagation, where its weights are initialized using a pre-trained CNN and GRU.
The proposed algorithm illustrates the state-of-the-art performance on all available public ImageQA benchmarks.
\vspace{-0.11cm}
\end{abstract}

\section{Introduction}
\label{sec:introduction}


One of the ultimate goals in computer vision is holistic scene understanding~\cite{holistic}, which requires a system to capture various kinds of information such as objects, actions, events, scene, atmosphere, and their relations in many different levels of semantics.
Although significant progress on various recognition tasks~\cite{Texture, Decaf, oquab2014learning, Vgg16, Googlenet, Deepface, Scenerecognition} has been made in recent years, these works focus only on solving relatively simple recognition problems in controlled settings, where each dataset consists of concepts with similar level of understanding (\eg object, scene, bird species, face identity, action, texture \etc).
There has been less efforts made on solving various recognition problems simultaneously, which is more complex and realistic, even though this is a crucial step toward holistic scene understanding.

\begin{figure}[t]
\centering
\adjincludegraphics[width=0.49\linewidth, trim={0 0 0 {.15\height}}, clip] {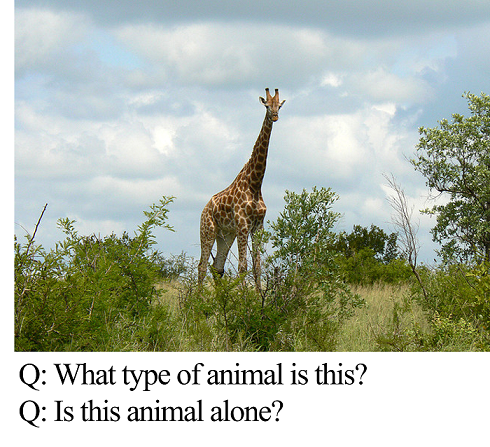}
\adjincludegraphics[width=0.49\linewidth, trim={0 0 0 {.15\height}}, clip] {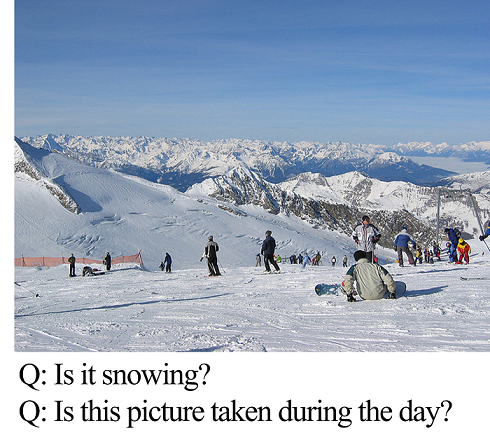}
 \\ 
\adjincludegraphics[width=0.49\linewidth, trim={0 0 0 {.15\height}}, clip] {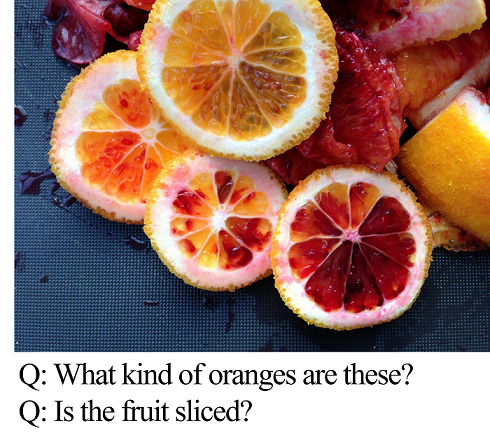}
\adjincludegraphics[width=0.49\linewidth, trim={0 0 0 {.15\height}}, clip] {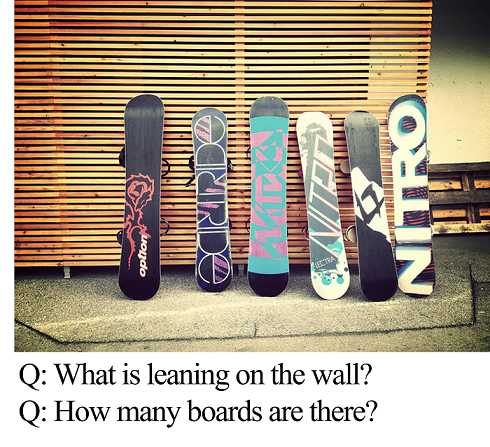}
\caption{Sample images and questions in VQA dataset~\cite{VQA}. Each question requires different type and/or level of understanding of the corresponding input image to find correct answers.}
\label{fig:iqa_eg}
\end{figure}


Image question answering (ImageQA)~\cite{VQA, Multiworld, mren2015} aims to solve the holistic scene understanding problem by proposing a task unifying various recognition problems.
ImageQA is a task automatically answering the questions about an input image as illustrated in Figure~\ref{fig:iqa_eg}.
The critical challenge of this problem is that different questions require different types and levels of understanding of an image to find correct answers.
For example, to answer the question like ``how is the weather?'' we need to perform classification on multiple choices related to weather, while we should decide between yes and no for the question like ``is this picture taken during the day?''
For this reason, not only the performance on a single recognition task but also the capability to select a proper task is important to solve ImageQA problem.


ImageQA problem has a short history in computer vision and machine learning community, but there already exist several approaches~\cite{Baiduqa, Convqa, Multiworld, Askneurons, mren2015}.
Among these methods, simple deep learning based approaches that perform classification on a combination of features extracted from image and question currently demonstrate the state-of-the-art accuracy on public benchmarks~\cite{mren2015, Convqa}; these approaches extract image features using a convolutional neural network (CNN), and use CNN or bag-of-words to obtain feature descriptors from question.
They can be interpreted as a method that the answer is given by the co-occurrence of a particular combination of features extracted from an image and a question.


Contrary to the existing approaches, we define a different recognition task depending on a question.
To realize this idea, we propose a deep CNN with a dynamic parameter layer whose weights are determined adaptively based on questions.
We claim that a single deep CNN architecture can take care of various tasks by allowing adaptive  weight assignment in the dynamic parameter layer.
For the adaptive parameter prediction, we employ a parameter prediction network, which consists of gated recurrent units (GRU) taking a question as its input and a fully-connected layer generating a set of candidate weights for the dynamic parameter layer.
The entire network including the CNN for ImageQA and the parameter prediction network is trained end-to-end through back-propagation, where its weights are initialized using pre-trained CNN and GRU.
Our main contributions in this work are summarized below:
\begin{itemize}
\item We successfully adopt a deep CNN with a dynamic parameter layer for ImageQA, which is a fully-connected layer whose parameters are determined dynamically based on a given question.
\item To predict a large number of weights in the dynamic parameter layer effectively and efficiently, we apply hashing trick~\cite{Hashing}, which reduces the number of parameters significantly with little impact on network capacity.
\item We fine-tune GRU pre-trained on a large-scale text corpus~\cite{Skipthought} to improve generalization performance of our network. Pre-training GRU on a large corpus is natural way to deal with a small number of training data, but no one has attempted it yet to our knowledge.
\item This is the first work to report the results on all currently available benchmark datasets such as DAQUAR, COCO-QA and VQA. 
Our algorithm achieves the state-of-the-art performance on all the three datasets.
\end{itemize}


The rest of this paper is organized as follows. 
We first review related work in Section~\ref{sec:related}. 
Section~\ref{sec:overview} and \ref{sec:architecture} describe the overview of our algorithm and the architecture of our network, respectively.
We discuss the detailed procedure to train the proposed network in Section~\ref{sec:training}.
Experimental results are demonstrated in Section~\ref{sec:experiment}.

\section{Related Work}
\label{sec:related}



There are several recent papers to address ImageQA~\cite{VQA, Baiduqa, Convqa, Multiworld, Askneurons, mren2015}; the most of them are based on deep learning except \cite{Multiworld}.
Malinowski and Fritz~\cite{Multiworld} propose a Bayesian framework, which exploits recent advances in computer vision and natural language processing. 
Specifically, it employs semantic image segmentation and symbolic question reasoning to solve ImageQA problem.
However, this method depends on a pre-defined set of predicates, which makes it difficult to represent complex models required to understand input images.


Deep learning based approaches demonstrate competitive performances in ImageQA~\cite{Askneurons, Baiduqa, mren2015, Convqa, VQA}.
Most approaches based on deep learning commonly use CNNs to extract features from image while they use different strategies to handle question sentences.
Some algorithms employ embedding of joint features based on image and question~\cite{VQA, Baiduqa, Askneurons}.
However, learning a softmax classifier on the simple joint features---concatenation of CNN-based image features and continuous bag-of-words representation of a question---performs better than LSTM-based embedding on COCO-QA~\cite{mren2015} dataset.
Another line of research is to utilize CNNs for feature extraction from both image and question and  combine the two features~\cite{Convqa}; this approach demonstrates impressive performance enhancement on DAQUAR~\cite{Multiworld} dataset by allowing fine-tuning the whole parameters.


The prediction of the weight parameters in deep neural networks has been explored in \cite{ba2015predicting} in the context of zero-shot learning.
To perform classification of unseen classes, it trains a multi-layer perceptron to predict a binary classifier for class-specific description in text.
However, this method is not directly applicable to ImageQA since finding solutions based on the combination of question and answer is a more complex problem than the one discussed in \cite{ba2015predicting}, and ImageQA involves a significantly larger set of candidate answers, which requires much more parameters than the binary classification case.
Recently, a parameter reduction technique based on a hashing trick is proposed by Chen~\etal~\cite{Hashing} to fit a large neural network in a limited memory budget. 
However, applying this technique to the dynamic prediction of parameters in deep neural networks is not attempted yet to our knowledge. 




\section{Algorithm Overview}
\label{sec:overview}
We briefly describe the motivation and formulation of our approach in this section.

\begin{figure*}
\centering
\includegraphics[width=0.95\linewidth] {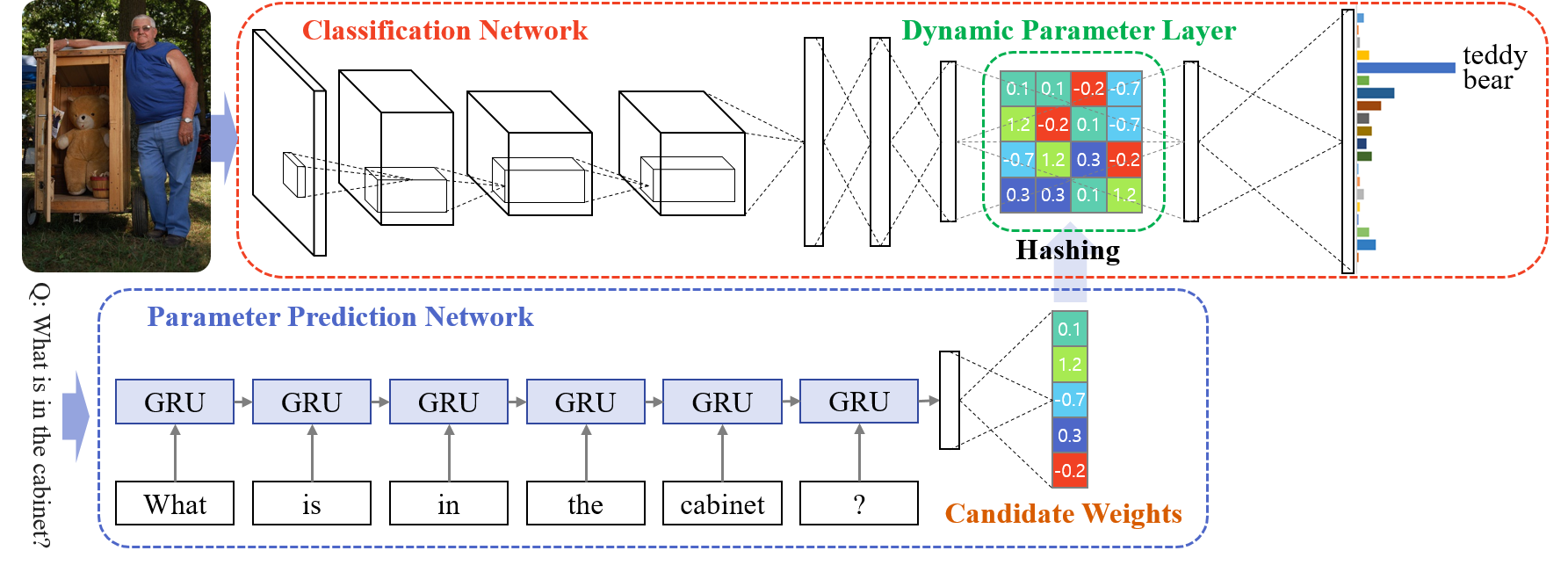}
\caption{Overall architecture of the proposed Dynamic Parameter Prediction network (DPPnet), which is composed of the classification network and the parameter prediction network. The weights in the dynamic parameter layer are mapped by a hashing trick from the candidate weights obtained from the parameter prediction network.
}
\label{fig:overall}
\end{figure*}


\subsection{Motivation}
\label{sub:motivation}


Although ImageQA requires different types and levels of image understanding, existing approaches~\cite{VQA, Baiduqa, Askneurons} pose the problem as a flat classification task.
However, we believe that it is difficult to solve ImageQA using a single deep neural network with fixed parameters.
In many CNN-based recognition problems, it is well-known to fine-tune a few layers for the adaptation to new tasks.
In addition, some networks are designed to solve two or more tasks jointly by constructing multiple branches connected to a common CNN architecture.
In this work, we hope to solve the heterogeneous recognition tasks using a single CNN by adapting the weights in the dynamic parameter layer.
Since the task is defined by the question in ImageQA, the weights in the layer are determined depending on the question sentence.
In addition, a hashing trick is employed to predict a large number of weights in the dynamic parameter layer and avoid parameter explosion.

\subsection{Problem Formulation}

ImageQA systems predict the best answer $\hat{a}$ given an image $I$ and a question $q$.
Conventional approaches~\cite{Convqa, mren2015} typically construct a joint feature vector based on two inputs $I$ and $q$ and solve a classification problem for ImageQA using the following equation:
\begin{equation}
\hat{a} = \underset{a\in{\Omega}}{\operatorname{argmax}} \hspace{0.1cm} p(a \vert {I}, q;{\bm{\theta}})
\end{equation}
where $\Omega$ is a set of all possible answers and ${\bm{\theta}}$ is a vector for the parameters in the network. 
On the contrary, we use the question to predict weights in the classifier and solve the problem.
We find the solution by 
\begin{equation}
\hat{a} = \underset{a\in{\Omega}}{\operatorname{argmax}} \hspace{0.1cm} p(a \vert I ;{\bm{\theta}}_{s},{\bm{\theta}}_{d} (q))
\end{equation}
where ${\bm{\theta}}_{s}$ and ${\bm{\theta}}_{d}(q)$ denote static and dynamic parameters, respectively.
Note that the values of ${\bm{\theta}}_{d}(q)$ are determined by the question $q$.

\section{Network Architecture}
\label{sec:architecture}

Figure~\ref{fig:overall} illustrates the overall architecture of the proposed algorithm.
The network is composed of two sub-networks: classification network and parameter prediction network.
The classification network is a CNN.
One of the fully-connected layers in the CNN is the dynamic parameter layer, and the weights in the layer are determined adaptively by the parameter prediction network.
The parameter prediction network has GRU cells and a fully-connected layer.
It takes a question as its input, and generates a real-valued vector, which corresponds to candidate weights for the dynamic parameter layer in the classification network.
Given an image and a question, our algorithm estimates the weights in the dynamic parameter layer through hashing with the candidate weights obtained from the parameter prediction network.
Then, it feeds the input image to the classification network to obtain the final answer.
More details of the proposed network are discussed in the following subsections.

\subsection{Classification Network}


The classification network is constructed based on VGG 16-layer net~\cite{Vgg16}, which is pre-trained on ImageNet~\cite{ImageNet}.
We remove the last layer in the network and attach three fully-connected layers.
The second last fully-connected layer of the network is the dynamic parameter layer whose weights are determined by the parameter prediction network, and the last fully-connected layer is the classification layer whose output dimensionality is equal to the number of possible answers.
The probability for each answer is computed by applying a softmax function to the output vector of the final layer.


We put the dynamic parameter layer in the second last fully-connected layer instead of the classification layer because it involves the smallest number of parameters.
As the number of parameters in the classification layer increases in proportion to the number of possible answers, predicting the weights for the classification layer may not be a good option to general ImageQA problems in terms of scalability.
Our choice for the dynamic parameter layer can be interpreted as follows. 
By fixing the classification layer while adapting the immediately preceding layer, we obtain the task-independent semantic embedding of all possible answers and use the representation of an input embedded in the answer space to solve an ImageQA problem.
Therefore, the relationships of the answers globally learned from all recognition tasks can help solve new ones involving unseen classes, especially in multiple choice questions.
For example, when not the exact ground-truth word (\eg, kitten) but similar words (\eg, cat and kitty) are shown at training time, the network can still predict the close answers (\eg, kitten) based on the globally learned answer embedding.
Even though we could also exploit the benefit of answer embedding based on the relations among answers to define a loss function, we leave it as our future work.

\subsection{Parameter Prediction Network}
\label{sub:paramter}


As mentioned earlier, our classification network has a dynamic parameter layer.
That is, for an input vector of the dynamic parameter layer ${\bf{f}}^{i}=\left[f^{i}_1,\dots,f^{i}_{\scriptscriptstyle N}\right]^{\scriptscriptstyle T}$, its output vector denoted by ${\bf{f}}^{o}=\left[f^{o}_1,\dots,f^{o}_{\scriptscriptstyle M}\right]^{\scriptscriptstyle T}$ is given by
\begin{equation}
{\bf{f}}^{o} = {\bf{W}}_{d}(q){\bf{f}}^{i}+{\bf{b}}
\end{equation}
where ${\bf{b}}$ denotes a bias and ${\bf{W}}_{d}(q)\in\mathbb{R}^{\scriptscriptstyle M \times N}$ denotes the matrix constructed dynamically using the parameter prediction network given the input question.
In other words, the weight matrix corresponding to the layer is parametrized by a function of the input question $q$.



\begin{figure}[t]
\centering
\includegraphics[width=1\linewidth] {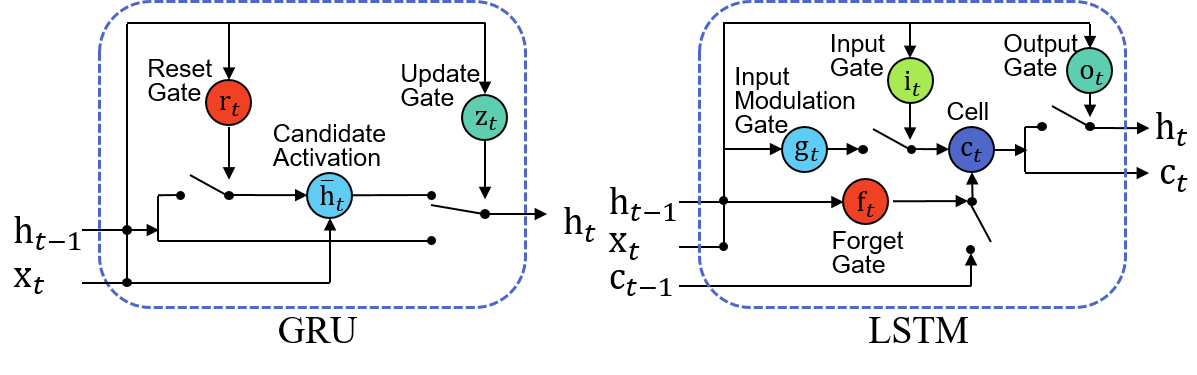}
\caption{Comparison of GRU and LSTM. Contrary to LSTM that contains memory cell explicitly, GRU updates the hidden state directly.}
\label{fig:gru_lstm}
\end{figure}

The parameter prediction network is composed of GRU cells~\cite{chung2014empirical} followed by a fully-connected layer, which produces the candidate weights to be used for the construction of weight matrix in the dynamic parameter layer within the classification network.
GRU, which is similar to LSTM, is designed to model dependency in multiple time scales.
As illustrated in Figure~\ref{fig:gru_lstm}, such dependency is captured by adaptively updating its hidden states with gate units.
However, contrary to LSTM, which maintains a separate memory cell explicitly, GRU directly updates its hidden states with a reset gate and an update gate. 
The detailed procedure of the update is described below.

Let $w_{1} ,...,w_{\scriptscriptstyle T}$ be the words in a question $q$, where $T$ is the number of words in the question.
In each time step $t$, given the embedded vector ${\bf{x}}_t$ for a word $w_t$, the GRU encoder updates its hidden state at time $t$, denoted by ${\bf h}_t$, using the following equations:
\begin{align}
{\bf{r}}_{t} &= \sigma({\bf{W}}_r{\bf{x}}_{t}+{\bf U}_r{\bf{h}}_{t-1}) \\
{\bf{z}}_{t} &= \sigma({\bf{W}}_z{\bf{x}}_{t}+{\bf U}_z{\bf{h}}_{t-1}) \\
\bar{{\bf{h}}}_{t} &= \tanh({\bf{W}}_h{\bf{x}}_{t} + {\bf U}_h({\bf{r}}_{t}\odot{\bf{h}}_{t-1})) \\
{\bf{h}}_{t} &= (1-{\bf{z}}_{t})\odot{\bf{h}}_{t-1}+{\bf{z}}_t\odot\bar{{\bf{h}}}_{t} 
\end{align}
where ${\bf{r}}_{t}$ and ${\bf{z}}_{t}$ respectively denote the reset and update gates at time ${t}$, and $\bar{{\bf{h}}}_{t}$ is candidate activation at time $t$.
In addition, $\odot$ indicates element-wise multiplication operator and $\sigma(\cdot)$ is a sigmoid function.
Note that the coefficient matrices related to GRU such as ${\bf{W}}_r$, ${\bf{W}}_z$, ${\bf{W}}_h$, ${\bf{U}}_r$, ${\bf{U}}_z$, and ${\bf{U}}_h$ are learned by our training algorithm.
By applying this encoder to a question sentence through a series of GRU cells, we obtain the final embedding vector ${\bf{h}}_{\scriptscriptstyle T}\in\mathbb{R}^{\scriptscriptstyle L}$ of the question sentence.


Once the question embedding is obtained by GRU, the candidate weight vector, ${\bf{p}}=\left[p_{1},\dots,p_{\scriptscriptstyle K}\right]^{\rm T}$, is given by applying a fully-connected layer to the embedded question ${\bf{h}}_{\scriptscriptstyle T}$ as
\begin{equation}
{\bf{p}} = {\bf{W}}_{p}{\bf{h}}_{\scriptscriptstyle T}
\end{equation}
where ${\bf{p}} \in \mathbb{R}^{\scriptscriptstyle K}$ is the output of the parameter prediction network, and ${\bf{W}}_{p}$ is the weight matrix of the fully-connected layer in the parameter prediction network.
Note that even though we employ GRU for a parameter prediction network since the pre-trained network for sentence embedding---skip-thought vector model~\cite{Skipthought}---is based on GRU, any form of neural networks, \eg, fully-connected and convolutional neural network, can be used to construct the parameter prediction network.

\subsection{Parameter Hashing}
\label{sub:hashing}


The weights in the dynamic parameter layers are determined based on the learned model in the parameter prediction network given a question.
The most straightforward approach to obtain the weights is to generate the whole matrix ${\bf{W}}_{d}(q)$ using the parameter prediction network.
However, the size of the matrix is very large, and the network may be overfitted easily given the limited number of training examples.
In addition, since we need quadratically more parameters between GRU and the fully-connected layer in the parameter prediction network to increase the dimensionality of its output, it is not desirable to predict full weight matrix using the network.
Therefore, it is preferable to construct ${\bf{W}}_{d}(q)$ based on a small number of candidate weights using a hashing trick.


We employ the recently proposed random weight sharing technique based on hashing~\cite{Hashing} to construct the weights in the dynamic parameter layer. 
Specifically, a single parameter in the candidate weight vector ${\bf{p}}$ is shared by multiple elements of ${\bf{W}}_{d}(q)$, which is done by applying a predefined hash function that converts the 2D location in ${\bf{W}}_{d}(q)$ to the 1D index in ${\bf{p}}$.
By this simple hashing trick, we can reduce the number of parameters in ${\bf{W}}_{d}(q)$ while maintaining the accuracy of the network~\cite{Hashing}.


Let $w^{d}_{mn}$ be the element at $(m,n)$ in ${\bf{W}}_{d}(q)$, which corresponds to the weight between $m^{\rm th}$ output and $n^{\rm th}$ input neuron. 
Denote by $\psi(m,n)$ a hash function mapping a key $(m,n)$ to a natural number in $\left\{1,\dots,K \right\}$, where $K$ is the dimensionality of ${\bf p}$.
The final hash function is given by
\begin{equation}
w^{d}_{mn}={p}_{\psi(m,n)} \cdot \xi(m,n)
\end{equation}
where $\xi(m,n):\mathbb{N}\times\mathbb{N}\rightarrow \{+1, -1 \}$ is another hash function independent of $\psi(m,n)$.
This function is useful to remove the bias of hashed inner product~\cite{Hashing}.
In our implementation of the hash function, we adopt an open-source implementation of {\it{xxHash}}\footnote{\url{https://code.google.com/p/xxhash/}}.


We believe that it is reasonable to reduce the number of free parameters based on the hashing technique as there are many redundant parameters in deep neural networks~\cite{denil2013predicting} and the network can be parametrized using a smaller set of candidate weights.
Instead of training a huge number of parameters without any constraint, it would be advantageous practically to allow multiple elements in the weight matrix to share the same value.
It is also demonstrated that the number of free parameter can be reduced substantially with little loss of network performance~\cite{Hashing}.

\section{Training Algorithm}
\label{sec:training}

This section discusses the error back-propagation algorithm in the proposed network and introduces the techniques adopted to enhance performance of the network.

\subsection{Training by Error Back-Propagation}


The proposed network is trained end-to-end to minimize the error between the ground-truths and the estimated answers.
The error is back-propagated by chain rule through both the classification network and the parameter prediction network and they are jointly trained by a first-order optimization method.


Let ${\mathcal{L}}$ denote the loss function.
The partial derivatives of ${\mathcal{L}}$ with respect to the $k^{\rm th}$ element in the input and output of the dynamic parameter layer are given respectively by 
\begin{equation}
{\delta}^{i}_k \equiv \frac{\partial\mathcal{L}}{\partial {f}^{i}_k}  ~~~~\text{and}~~~~
{\delta}^{o}_k \equiv \frac{\partial\mathcal{L}}{\partial {f}^{o}_k}.
\end{equation}
The two derivatives have the following relation:
\begin{equation}
{\delta}^{i}_n = \sum _{ m=1 }^{ M }{ w^d_{mn}\delta^o_{m} } 
\end{equation}
Likewise, the derivative with respect to the assigned weights in the dynamic parameter layer is given by
\begin{equation}
{\frac{\partial\mathcal{L}}{\partial w^{d}_{mn}}}=f^{i}_{n}{\delta}^{o}_{m}.
\end{equation}
%
As a single output value of the parameter prediction network is shared by multiple connections in the dynamic parameter layer, the derivatives with respect to all shared weights need to be accumulated to compute the derivative with respect to an element in the output of the parameter prediction network as follows:
\begin{align}
{\frac{\partial\mathcal{L}}{\partial p_{k}}}
&= \sum _{m=1}^{\scriptscriptstyle M}{\sum _{n=1}^{\scriptscriptstyle N}{   {\frac{\partial\mathcal{L}}{\partial w^d_{mn}}} {\frac{\partial w^d_{mn}}{\partial p_{k}}}}}  \nonumber \\
&= \sum _{m=1}^{\scriptscriptstyle M}{\sum _{n=1}^{\scriptscriptstyle N} {  {\frac{\partial\mathcal{L}}{\partial w^d_{mn}}}   {\xi(m,n)}  {{\mathbb{I}} [ \psi(m,n)=k ]}      }},
\end{align}
where ${\mathbb{I}} [ \cdot ]$ denotes the indicator function.
The gradients of all the preceding layers in the classification and parameter prediction networks are computed by the standard back-propagation algorithm.

\subsection{Using Pre-trained GRU}

Although encoders based on recurrent neural networks (RNNs) such as LSTM~\cite{LSTM} and GRU~\cite{chung2014empirical} demonstrate impressive performance on sentence embedding~\cite{mikolov2010recurrent, sutskever2014sequence}, their benefits in the ImageQA task are marginal in comparison to bag-of-words model~\cite{mren2015}.
One of the reasons for this fact is the lack of language data in ImageQA dataset.
Contrary to the tasks that have large-scale training corpora, even the largest ImageQA dataset contains relatively small amount of language data; for example, \cite{VQA} contains 750K questions in total.
Note that the model in \cite{sutskever2014sequence} is trained using a corpus with more than 12M sentences.

To deal with the deficiency of linguistic information in ImageQA problem, we transfer the information acquired from a large language corpus by fine-tuning the pre-trained embedding network.
We initialize the GRU with the skip-thought vector model trained on a book-collection corpus containing more than 74M sentences~\cite{Skipthought}.
Note that the GRU of the skip-thought vector model is trained in an unsupervised manner by predicting the surrounding sentences from the embedded sentences.
As this task requires to understand context, the pre-trained model produces a generic sentence embedding, which is difficult to be trained with a limited number of training examples.
By fine-tuning our GRU initialized with a generic sentence embedding model for ImageQA, we obtain the representations for questions that are generalized better.


\subsection{Fine-tuning CNN}


It is very common to transfer CNNs for new tasks in classification problems, but it is not trivial to fine-tune the CNN in our problem.
We observe that the gradients below the dynamic parameter layer in the CNN are noisy since the weights are predicted by the parameter prediction network.
Hence, a straightforward approach to fine-tune the CNN typically fails to improve performance, and we employ a slightly different technique for CNN fine-tuning to sidestep the observed problem. 
We update the parameters of the network using new datasets except the part transferred from VGG 16-layer net at the beginning, and start to update the weights in the subnetwork if the validation accuracy is saturated.

\subsection{Training Details}


Before training, question sentences are normalized to lower cases and preprocessed by a simple tokenization technique as in \cite{Showattend}.
We normalize the answers to lower cases and regard a whole answer in a single or multiple words as a separate class.

The network is trained end-to-end by back-propagation.
Adam~\cite{Adam} is used for optimization with initial learning rate 0.01.
We clip the gradient to 0.1 to handle the gradient explosion from the recurrent structure of GRU~\cite{pascanu2013difficulty}.
Training is terminated when there is no progress on validation accuracy for 5 epochs.

Optimizing the dynamic parameter layer is not straightforward since the distribution of the outputs in the dynamic parameter layer is likely to change significantly in each batch.
Therefore, we apply batch-normalization~\cite{Batchnorm} to the output activations of the layer to alleviate this problem.
%
In addition, we observe that GRU tends to converge fast and overfit data easily if training continues without any restriction.
We stop fine-tuning GRU when the network start to overfit and continue to train the other parts of the network; this strategy improves performance in practice.

\section{Experiments}
\label{sec:experiment}
We now describe the details of our implementation and evaluate the proposed method in various aspects.

\subsection{Datasets}
We evaluate the proposed network on all public ImageQA benchmark datasets such as DAQUAR~\cite{Multiworld}, COCO-QA~\cite{mren2015} and VQA~\cite{VQA}. 
They collected question-answer pairs from existing image datasets and most of the answers are single words or short phrases.

DAQUAR is based on NYUDv2~\cite{Nyud} dataset, which is originally designed for indoor segmentation using RGBD images.
DAQUAR provides two benchmarks, which are distinguished by the number of classes and the amount of data; DAQUAR-all consists of 6,795 and 5,673 questions for training and testing respectively, and includes 894 categories in answer.
DAQUAR-reduced includes only 37 answer categories for 3,876 training and 297 testing questions.
Some questions in this dataset are associated with a set of multiple answers instead of a single one.

The questions in COCO-QA are automatically generated from the image descriptions in MS COCO dataset~\cite{Mscoco} using the constituency parser with simple question-answer generation rules.
The questions in this dataset are typically long and explicitly classified into 4 types depending on the generation rules: object questions, number questions, color questions and location questions.
All answers are with one-words and there are 78,736 questions for training and 38,948 questions for testing.

Similar to COCO-QA, VQA is also constructed on MS COCO~\cite{Mscoco} but each question is associated with multiple answers annotated by different people. 
This dataset contains the largest number of questions: 248,349 for training, 121,512 for validation, and 244,302 for testing, where the testing data is splited into test-dev, test-standard, test-challenge and test-reserve as in~\cite{Mscoco}.
Each question is provided with 10 answers to take the consensus of annotators into account.
About 90\% of answers have single words and 98\% of answers do not exceed three words.

\subsection{Evaluation Metrics}

DAQUAR and COCO-QA employ both classification accuracy and its relaxed version based on word similarity, WUPS~\cite{Multiworld}.
It uses thresholded Wu-Palmer similarity~\cite{wu1994verbs} based on WordNet~\cite{fellbaum1998wordnet} taxonomy to compute the similarity between words.
For predicted answer set $\mathcal{A}^{i}$ and ground-truth answer set $\mathcal{T}^{i}$ of the $i^{\rm th}$ example, WUPS is given by
\begin{align}
&{\text{WUPS}} = \nonumber \\ 
&\frac { 1 }{ N } \sum _{ i=1 }^{ N }{ \min { \left\{ \prod _{ a\in \mathcal{A}^{ i } }^{  }{ \max _{ t\in \mathcal{T}^{ i } }{ \mu \left( a,t \right)  }  } , \prod _{ t\in \mathcal{T}^{ i } }^{  }{ \max _{ a\in \mathcal{A}^i }{ \mu \left( a,t \right)  }  }  \right\}  }  } , 
\end{align}
where $\mu \left(\cdot, \cdot \right)$ denotes the thresholded Wu-Palmer similarity between prediction and ground-truth.
We use two threshold values ($0.9$ and $0.0$) in our evaluation.

VQA dataset provides open-ended task and multiple-choice task for evaluation.
For open-ended task, the answer can be any word or phrase while an answer should be chosen out of 18 candidate answers in the multiple-choice task.
In both cases, answers are evaluated by accuracy reflecting human consensus.
For predicted answer $a_i$ and target answer set $\mathcal{T}^{i}$ of the $i^{\rm th}$ example, the accuracy is given by
\begin{equation}
\text{Acc}_\textrm{VQA} = \frac{1}{N} \sum _{i=1}^{N}{\min { \left\{ \frac{\sum _{t\in \mathcal{T}^{i}}{  {\mathbb{I}} \left[a_i=t\right]}}{3} , 1\right\} }  }
\end{equation}
where ${\mathbb{I}}\left[ \cdot \right]$ denotes an indicator function.
In other words, a predicted answer is regarded as a correct one if at least three annotators agree, and the score depends on the number of agreements if the predicted answer is not correct.

\subsection{Results}

We test three independent datasets, VQA, COCO-QA, and DAQUAR, and first present the results for VQA dataset in Table~\ref{tab:vqa_result}.
The proposed Dynamic Parameter Prediction network (DPPnet) outperforms all existing methods nontrivially.
We performed controlled experiments to analyze the contribution of individual components in the proposed algorithm---dynamic parameter prediction, use of pre-trained GRU and CNN fine-tuning, and trained 3 additional models, CONCAT, RAND-GRU, and CNN-FIXED.
CNN-FIXED is useful to see the impact of CNN fine-tuning since it is identical to DPPnet except that the weights in CNN are fixed.
RAND-GRU is the model without GRU pre-training, where the weights of GRU and word embedding model are initialized randomly.
It does not fine-tune CNN either.
CONCAT is the most basic model, which predicts answers using the two fully-connected layers for a combination of CNN and GRU features.
Obviously, it does not employ any of new components such as parameter prediction, pre-trained GRU and CNN fine-tuning.

\begin{table}[!t] \footnotesize
\centering
\caption{Evaluation results on VQA test-dev in terms of $\text{Acc}_\text{VQA}$} \vspace{0.1cm}
\begin{tabular}
{
@{}C{1.7cm}@{}|@{}C{0.8cm}@{}@{}C{0.8cm}@{}@{}C{0.9cm}@{}@{}C{0.8cm}@{}|@{}C{0.8cm}@{}@{}C{0.8cm}@{}@{}C{0.9cm}@{}@{}C{0.8cm}@{}
}
&\multicolumn{4}{c|}{Open-Ended} & \multicolumn{4}{c}{Multiple-Choice} \\
& All &Y/N & Num & Others & All &Y/N & Num & Others \\
\hline
Question~\cite{VQA}&48.09&75.66&36.70&27.14&53.68&75.71&37.05&38.64\\
Image~\cite{VQA}&28.13&64.01&00.42&03.77&30.53&69.87&00.45&03.76\\
Q+I~\cite{VQA}&52.64&75.55&33.67&37.37&58.97&75.59&34.35&50.33\\
LSTM Q~\cite{VQA}&48.76&78.20&35.68&26.59&54.75&78.22&36.82&38.78\\
LSTM Q+I~\cite{VQA}& 53.74&78.94&35.24&36.42&57.17&78.95&35.80&43.41\\
\hline
CONCAT&54.70&77.09&36.62&39.67&59.92&77.10&37.48&50.31\\ 
RAND-GRU&55.46&79.58&36.20&39.23&61.18&79.64&38.07&50.63\\ 
CNN-FIXED&56.74&80.48&37.20&40.90&61.95&80.56&38.32&51.40\\ 
DPPnet&{\bf{57.22}}&{\bf{80.71}}&{\bf{37.24}}&{\bf{41.69}}&{\bf{62.48}}&{\bf{80.79}}&{\bf{38.94}}&{\bf{52.16}}\\
\hline
\end{tabular}
\label{tab:vqa_result}
\end{table}
\begin{table}[!t] \footnotesize
\centering
\caption{Evaluation results on VQA test-standard} \vspace{0.1cm}
\begin{tabular}
{
@{}C{1.7cm}@{}|@{}C{0.8cm}@{}@{}C{0.8cm}@{}@{}C{0.9cm}@{}@{}C{0.8cm}@{}|@{}C{0.8cm}@{}@{}C{0.8cm}@{}@{}C{0.9cm}@{}@{}C{0.8cm}@{}
}
&\multicolumn{4}{c|}{Open-Ended} & \multicolumn{4}{c}{Multiple-Choice} \\
& All &Y/N & Num & Others & All &Y/N & Num & Others \\
\hline
Human~\cite{VQA}&83.30&95.77&83.39&72.67&-&-&-&-\\ 
\hline
LSTM Q+I~\cite{VQA}&54.06&-&-&-&-&-&-&-\\ 
\hline
DPPnet&{\bf{57.36}}&{\bf{80.28}}&{\bf{36.92}}&{\bf{42.24}}&{\bf{62.69}}&{\bf{80.35}}&{\bf{38.79}}&{\bf{52.79}}\\
\hline
\end{tabular}
\label{tab:vqa_result_std}
\end{table}

The results of the controlled experiment are also illustrated in Table~\ref{tab:vqa_result}.
CONCAT already outperforms LSTM Q+I by integrating GRU instead of LSTM~\cite{chung2014empirical} and batch normalization. 
RAND-GRU achieves better accuracy by employing dynamic parameter prediction additionally.
It is interesting that most of the improvement comes from yes/no questions, which may involve various kinds of tasks since it is easy to ask many different aspects in an input image for binary classification.
CNN-FIXED improves accuracy further by adding GRU pre-training, and our final model DPPnet achieves the state-of-the-art performance on VQA dataset with large margins as illustrated in Table~\ref{tab:vqa_result} and ~\ref{tab:vqa_result_std}. 

\begin{table}[!t] \footnotesize
\centering
\caption{Evaluation results on COCO-QA} \vspace{0.1cm}
\begin{tabular}
{
@{}C{2.3cm}@{}|@{}C{1.8cm}@{}@{}C{1.8cm}@{}@{}C{1.8cm}@{}
}
&Acc&WUPS 0.9&WUPS 0.0\\
\hline
IMG+BOW~\cite{mren2015}&55.92&66.78&88.99\\
2VIS+BLSTM~\cite{mren2015}&55.09&65.34&88.64\\
Ensemble~\cite{mren2015}&57.84&67.90&89.52\\
ConvQA~\cite{Convqa}&54.95&65.36&88.58\\
\hline
DPPnet&{\bf{61.19}}&{\bf{70.84}}&{\bf{90.61}}\\
\hline
\end{tabular}
\label{tab:cocoqa_result}
\end{table}

\begin{table}[!t] \footnotesize
\centering
\caption{Evaluation results on DAQUAR reduced} \vspace{0.1cm}
\begin{tabular}
{
@{}C{2.3cm}@{}|@{}C{0.9cm}@{}@{}C{1.0cm}@{}@{}C{0.8cm}@{}|@{}C{0.9cm}@{}@{}C{1.0cm}@{}@{}C{0.8cm}@{}
}
&\multicolumn{3}{c|}{Single answer}& \multicolumn{3}{c}{Multiple answers}\\
&Acc& 0.9& 0.0&Acc& 0.9& 0.0\\
\hline
Multiworld~\cite{Multiworld}&-&-&-&12.73&18.10&51.47\\
Askneuron~\cite{Askneurons}&34.68&40.76&79.54&29.27&36.50&79.47\\
IMG+BOW~\cite{mren2015}&34.17&44.99&81.48&-&-&-\\
2VIS+BLSTM~\cite{mren2015}&35.78&46.83&82.15&-&-&-\\
Ensemble~\cite{mren2015}&36.94&48.15&82.68&-&-&-\\
ConvQA~\cite{Convqa}&39.66&44.86&83.06&38.72&44.19&79.52\\
\hline
DPPnet&{\bf{44.48}}&{\bf{49.56}}&{\bf{83.95}}&{\bf{44.44}}&{\bf{49.06}}&{\bf{82.57}}\\

\hline
\end{tabular}
\label{tab:daquar_reduced_result}
\end{table}

\begin{table}[!t] \footnotesize
\centering
\caption{Evaluation results on DAQUAR all} \vspace{0.1cm}
\begin{tabular}
{
@{}C{2.3cm}@{}|@{}C{0.9cm}@{}@{}C{1.0cm}@{}@{}C{0.8cm}@{}|@{}C{0.9cm}@{}@{}C{1.0cm}@{}@{}C{0.8cm}@{}
}

&\multicolumn{3}{c|}{Single answer}& \multicolumn{3}{c}{Multiple answers}\\
&Acc& 0.9& 0.0&Acc& 0.9& 0.0\\
\hline
Human~\cite{Multiworld}&-&-&-&50.20&50.82&67.27\\
\hline
Multiworld~\cite{Multiworld}&-&-&-&07.86&11.86&38.79\\
Askneuron~\cite{Askneurons}&19.43&25.28&62.00&17.49&23.28&57.76\\
ConvQA~\cite{Convqa}&23.40&29.59&62.95&20.69&25.89&55.48\\
\hline
DPPnet&{\bf{28.98}}&{\bf{34.80}}&{\bf{67.81}}&{\bf{25.60}}&{\bf{31.03}}&{\bf{60.77}}\\
\hline
\end{tabular}
\label{tab:daquar_all_result}
\end{table}

Table~\ref{tab:cocoqa_result}, \ref{tab:daquar_reduced_result}, and \ref{tab:daquar_all_result} illustrate the results by all algorithms including ours that have reported performance on COCO-QA, DAQUAR-reduced, DAQUAR-all datasets.
The proposed algorithm outperforms all existing approaches consistently in all benchmarks.
In Table~\ref{tab:daquar_reduced_result} and \ref{tab:daquar_all_result}, single answer and multiple answers denote the two subsets of questions divided by the number of ground-truth answers.
Also, the numbers (0.9 and 0.0) in the second rows are WUPS thresholds.

To understand how the parameter prediction network understand questions, we present several representative questions before and after fine-tuning GRU in a descending order based on their cosine similarities to the query question in Table~\ref{tab:gru_finetuning}.
The retrieved sentences are frequently determined by common subjective or objective words before fine-tuning while they rely more on the tasks to be solved after fine-tuning.

\begin{table*}[!t] \footnotesize
\centering
\caption{Retrieved sentences before and after fine-tuning GRU} \vspace{0.1cm}
\begin{tabular}
{
@{}C{2.4cm}@{}|@{}C{9cm}@{}|@{}C{5.5cm}@{}
}
Query question&{\color{blue}What body part} has most recently contacted {\color{red}the ball}?&Is the person {\color{blue}feeding} {\color{red}the birds}?\\
\hline \hline
\multirow{8}{*}{Before fine-tuning}&What shape is {\color{red} the ball}?&Is he {\color{blue}feeding} {\color{red}the birds}?\\
      &What colors are {\color{red}the ball}?&Is the reptile fighting {\color{red}the birds}?\\
      &What team has {\color{red}the ball}?&Does the elephant want to play with {\color{red}the birds}?\\
      &How many times has the girl hit {\color{red}the ball}?&What is the fence made of behind {\color{red}the birds}?\\
      &What number is on the women's Jersey closest to {\color{red}the ball}?&Where are the majority of {\color{red}the birds}?\\
      &What is unusual about {\color{red}the ball}?&What colors are {\color{red}the birds}?\\
      &What is the speed of {\color{red}the ball}?&Is this man {\color{blue}feeding} the pigeons?\\
\hline
\multirow{8}{*}{After fine-tuning}&{\color{blue}What body part} is the boy holding the bear by? & Is he {\color{blue}feeding} {\color{red}the birds}?\\
      & {\color{blue}What body part} is on the right side of this picture? &Is the person {\color{blue}feeding} the sheep?\\
      & {\color{blue}What human body part} is on the table? &Is the man {\color{blue}feeding} the pigeons?\\
      & {\color{blue}What body parts} appear to be touching? &Is she {\color{blue}feeding} the pigeons?\\
      & {\color{blue}What partial body parts} are in the foreground? &Is that the zookeeper {\color{blue}feeding} the giraffes?\\
      & {\color{blue}What part of the body} does the woman on the left have on the ramp? &Is the reptile fighting {\color{red}the birds}?\\
      & {\color{blue}Name a body part} that would not be visible if the woman's mouth was closed? & Does the elephant want to play with {\color{red}the birds}? \\
\hline
\end{tabular}
\label{tab:gru_finetuning}
\end{table*}
\begin{figure*}[t]
\centering
\subfigure[Result of the proposed algorithm on multiple questions for a single image]{
\begin{minipage}{1\textwidth}
\centering
\adjincludegraphics[width=0.25\linewidth, trim={0 0 0 {.05\height}}, clip] {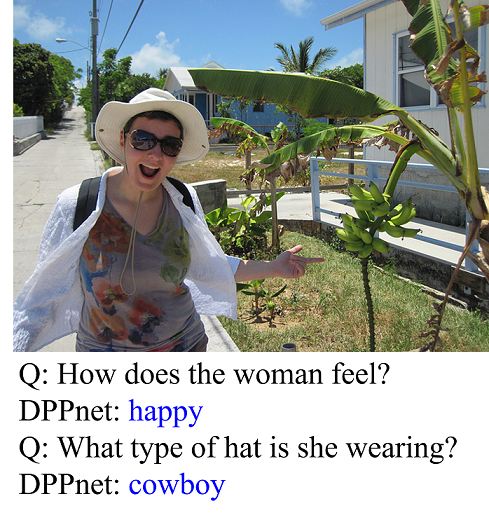}
\hspace{-0.2cm}
\adjincludegraphics[width=0.25\linewidth, trim={0 0 0 {.05\height}}, clip] {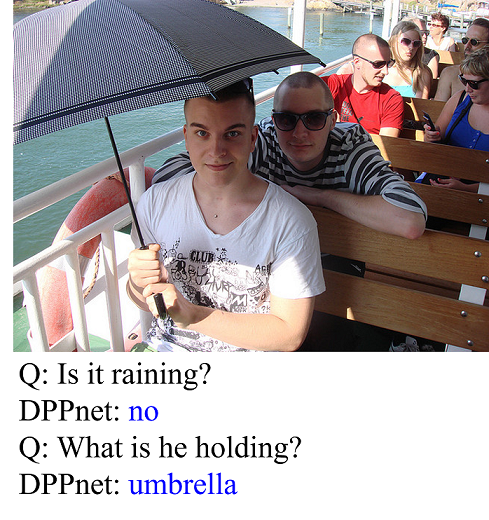}
\hspace{-0.2cm}
\adjincludegraphics[width=0.25\linewidth, trim={0 0 0 {.05\height}}, clip] {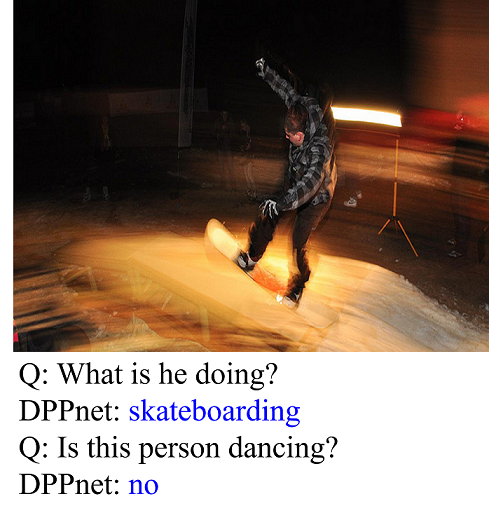}
\hspace{-0.2cm}
\adjincludegraphics[width=0.25\linewidth, trim={0 0 0 {.05\height}}, clip] {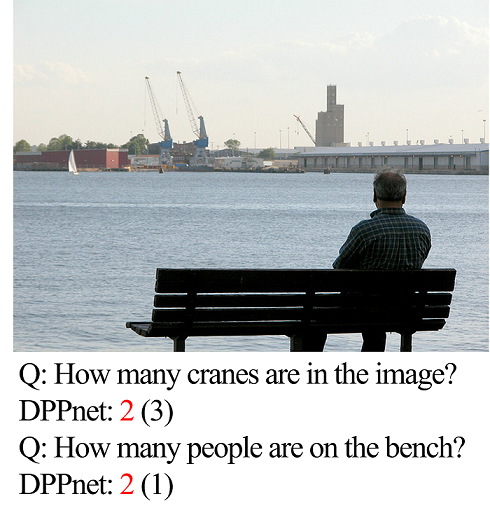}
 \\ 
\end{minipage}
\label{fig:qualitative_i1qn}
}\\
\vspace{-0.2cm}

\subfigure[Results of the proposed algorithm on a single common question for multiple images]{
\begin{minipage}{1\textwidth}
\centering
\includegraphics[width=0.50\linewidth] {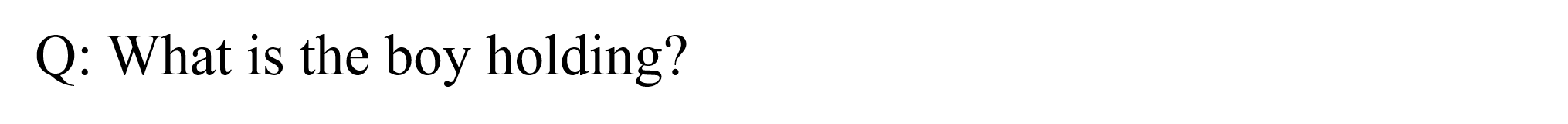}
\hspace{-0.2cm}
\includegraphics[width=0.50\linewidth] {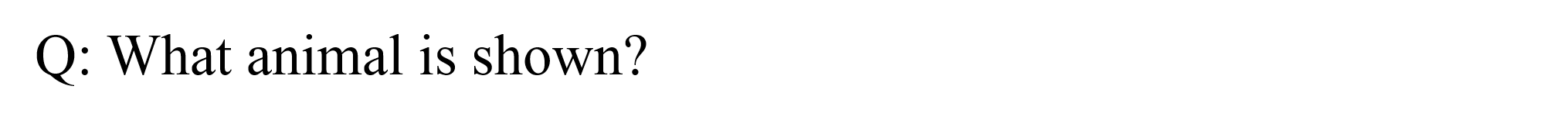}
 \\ \vspace{-0.2cm}
\adjincludegraphics[width=0.25\linewidth, trim={0 0 0 {.06\height}}, clip] {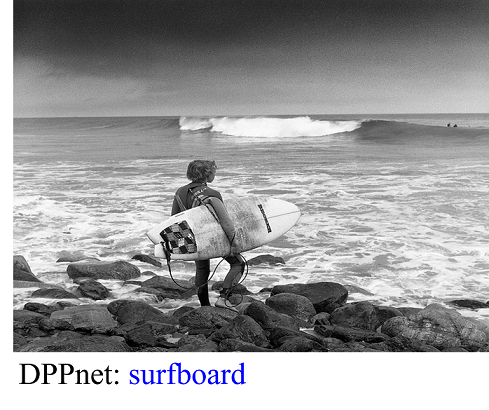}
\hspace{-0.2cm}
\adjincludegraphics[width=0.25\linewidth, trim={0 0 0 {.06\height}}, clip] {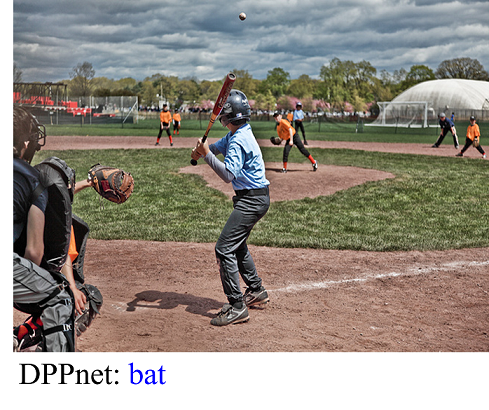}
\hspace{-0.2cm}
\adjincludegraphics[width=0.25\linewidth, trim={0 0 0 {.06\height}}, clip] {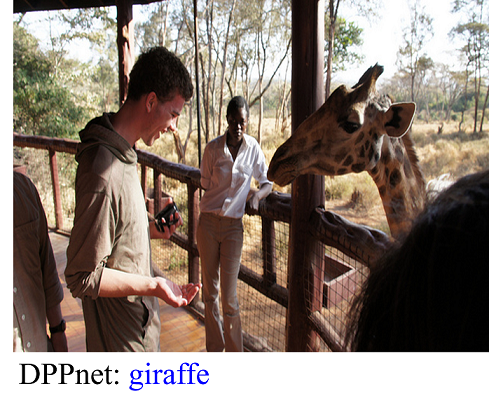}
\hspace{-0.2cm}
\adjincludegraphics[width=0.25\linewidth, trim={0 0 0 {.06\height}}, clip] {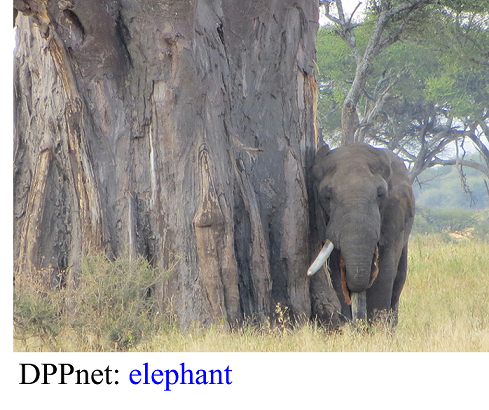}
 \\ \vspace{-0.2cm}
\includegraphics[width=0.50\linewidth] {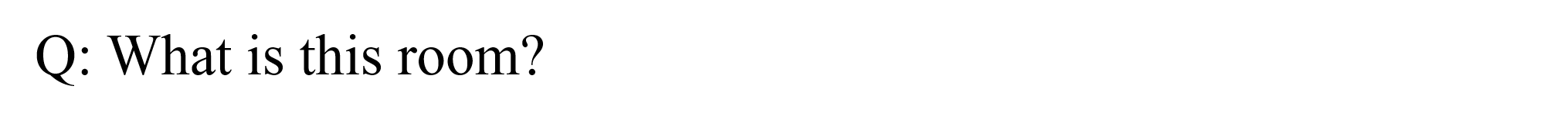}
\hspace{-0.2cm}
\includegraphics[width=0.50\linewidth] {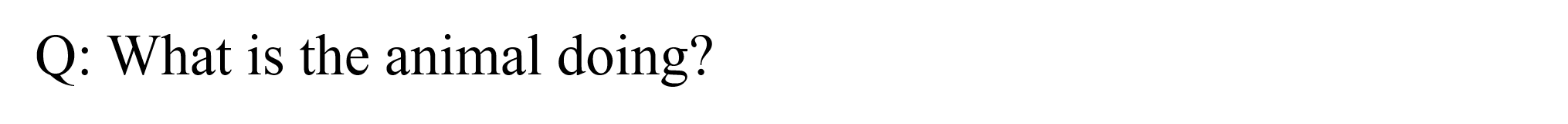}
 \\ \vspace{-0.2cm}
\adjincludegraphics[width=0.25\linewidth, trim={0 0 0 {.06\height}}, clip] {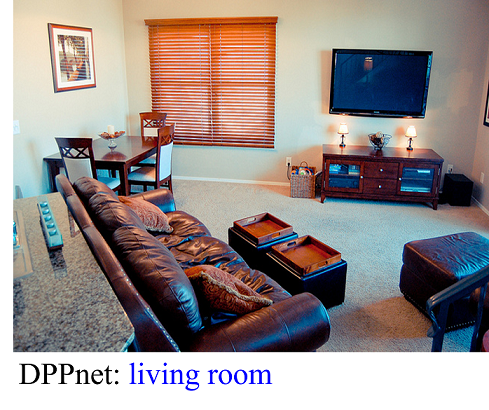}
\hspace{-0.2cm}
\adjincludegraphics[width=0.25\linewidth, trim={0 0 0 {.06\height}}, clip] {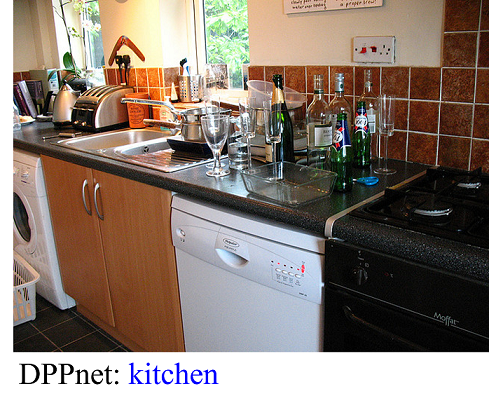}
\hspace{-0.2cm}
\adjincludegraphics[width=0.25\linewidth, trim={0 0 0 {.06\height}}, clip] {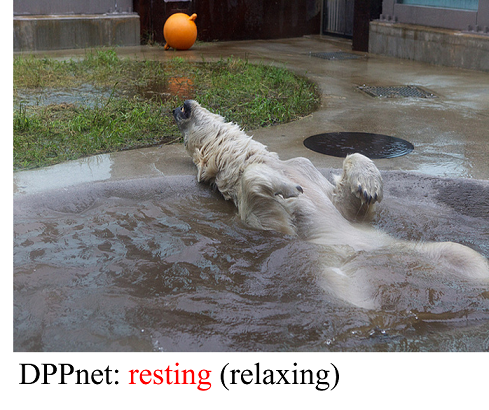}
\hspace{-0.2cm}
\adjincludegraphics[width=0.25\linewidth, trim={0 0 0 {.06\height}}, clip] {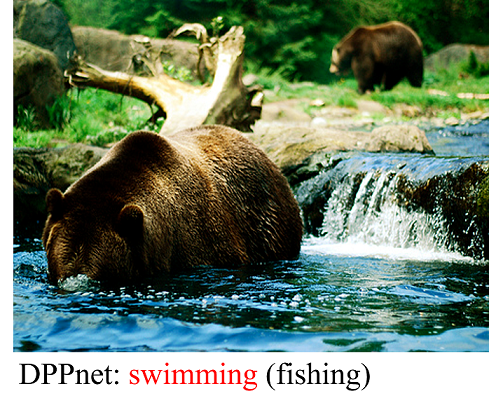}
\\
\end{minipage}
\label{fig:qualitative_q1in}
}\\

\caption{Sample images and questions in VQA dataset~\cite{VQA}. Each question requires a different type and/or level of understanding of the corresponding input image to find correct answer. Answers in blue are correct while answers in red are incorrect. For the incorrect answers,  ground-truth answers are provided within the parentheses.}
\label{fig:qualitative}
\end{figure*}

The qualitative results of the proposed algorithm are presented in Figure~\ref{fig:qualitative}.
In general, the proposed network is successful to handle various types of questions that need different levels of semantic understanding.
Figure~\ref{fig:qualitative_i1qn} shows that the network is able to adapt recognition tasks depending on questions.
However, it often fails in the questions asking the number of occurrences since these questions involve the difficult tasks (\eg, object detection) to learn only with image level annotations.
On the other hand, the proposed network is effective to find the answers for the same question on different images fairly well as illustrated in Figure~\ref{fig:qualitative_q1in}.
Refer to our project website\footnote{\url{http://cvlab.postech.ac.kr/research/dppnet/}} for more comprehensive qualitative results.

\section{Conclusion}
We proposed a novel architecture for image question answering based on two subnetworks---classification network and parameter prediction network.
The classification network has a dynamic parameter layer, which enables the classification network to adaptively determine its weights through the parameter prediction network.
While predicting all entries of the weight matrix is infeasible due to its large dimensionality, we relieved this limitation using parameter hashing and weight sharing.
The effectiveness of the proposed architecture is supported by experimental results showing the state-of-the-art performances on three different datasets.
Note that the proposed method achieved outstanding performance even without more complex recognition processes such as referencing objects.
We believe that the proposed algorithm can be extended further by integrating attention model~\cite{Showattend} to solve such difficult problems.

{\small
\bibliographystyle{ieee}
\bibliography{egbib}
}

\end{document}